\def\year{2020}\relax
\documentclass[letterpaper]{article} 
\usepackage{aaai20}  
\usepackage{times}  
\usepackage{helvet} 
\usepackage{courier}  
\usepackage[hyphens]{url}  
\usepackage{graphicx}  
\usepackage{xcolor}
\usepackage{soul}
\usepackage{amssymb}
\usepackage{url}
\usepackage[draft]{hyperref}
\usepackage[utf8]{inputenc}
\usepackage[small]{caption}
\usepackage{amsmath}
\usepackage{booktabs}
\usepackage{algorithm}
\usepackage{algorithmic}
\usepackage{bm}
\usepackage{footnote}

\title{Variational Student: Learning Compact and Sparser Networks in Knowledge Distillation Framework }
\author{Srinidhi Hegde\textsuperscript{\rm 1} \quad Ranjitha Prasad\textsuperscript{\rm 1} \quad Ramya Hebbalaguppe\textsuperscript{\rm 1} \quad Vishwajeet Kumar\textsuperscript{\rm 2} 
\\
\\
\textsuperscript{\rm 1} TCS Research, New Delhi \\
\textsuperscript{\rm 2} IIT Kharagpur
}
 
\begin{document}

\maketitle

\begin{abstract}
The holy grail in deep neural network research is porting the memory- and computation-intensive network models on embedded platforms with a minimal compromise in model accuracy. To this end, we propose a novel approach, termed as \emph{Variational Student}, where we reap the benefits of compressibility of the \textit{knowledge distillation} (KD) framework, and sparsity inducing abilities of \textit{variational inference} (VI) techniques. Essentially, we build a sparse student network, whose sparsity is induced by the variational parameters found via optimizing a loss function based on VI, leveraging the \emph{knowledge} learnt by an accurate but complex pre-trained teacher network. Further, for sparsity enhancement, we also employ a \textit{Block Sparse Regularizer} on a concatenated tensor of teacher and student network weights. We demonstrate that the marriage of KD and the VI techniques inherits compression properties from the KD framework, and enhances levels of sparsity from the VI approach, with minimal compromise in the model accuracy. We benchmark our results on LeNet $300-100$ (MLP) and VGGNet (CNN) and illustrate a memory footprint reduction of $\sim 64\times$ and $\sim 213\times$ on these MLP and CNN variants, respectively, without a need to retrain the teacher network. Furthermore, in the low data regime, we observed that our method outperforms state-of-the-art Bayesian techniques in terms of accuracy.
\end{abstract}

\section{Introduction}
\label{intro}

\begin{figure*}[]
	\begin{center}
		\includegraphics[scale=.4]{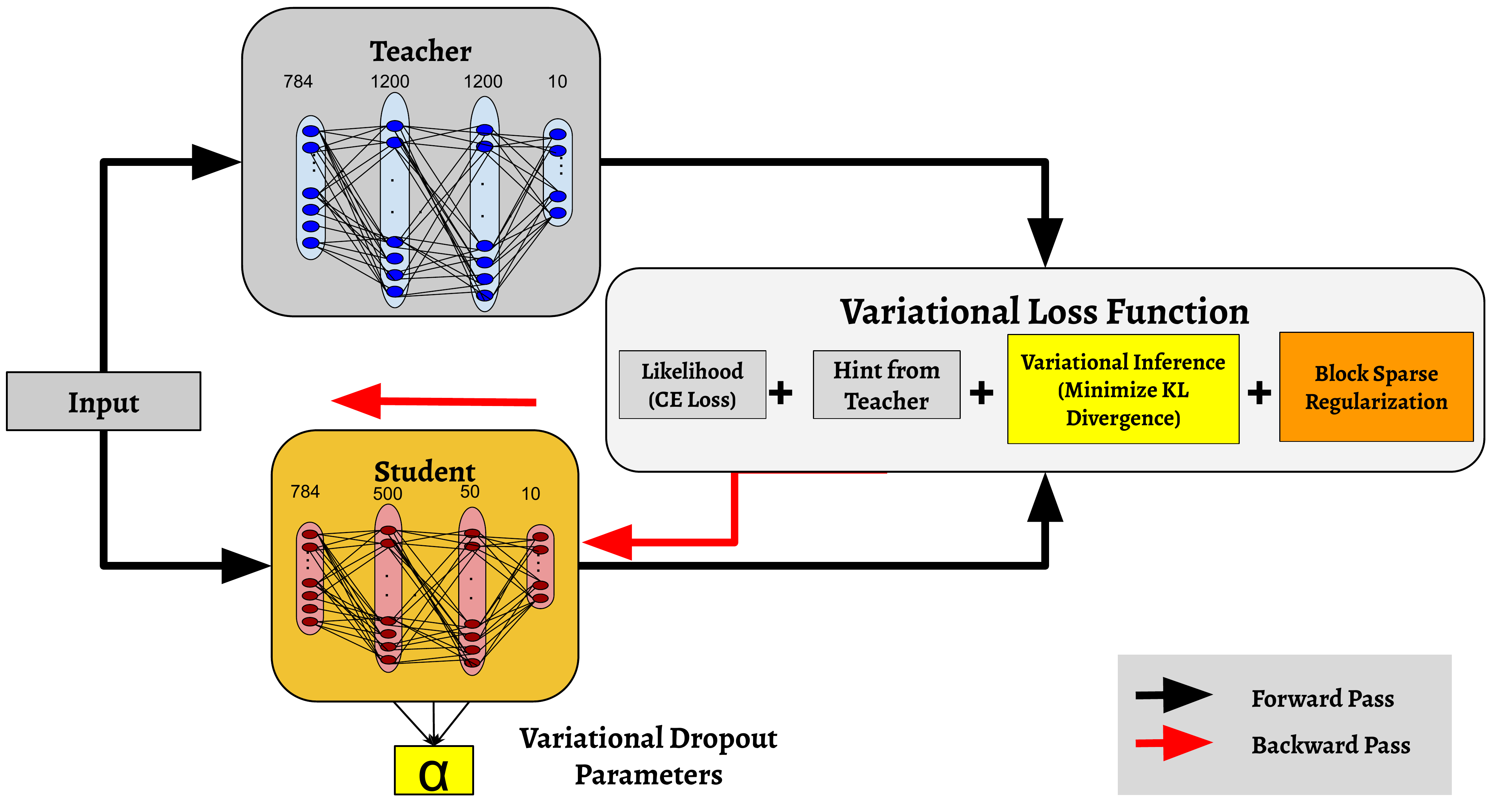}
	\end{center}
		\caption{Training procedure for learning compact and sparse student networks. The roles of different terms in variational loss function are: \textit{likelihood} -  for independent student network's learning; \textit{hint} - learning induced from teacher network; \textit{variational term} -  promotes sparsity by optimizing variational dropout parameters, $\alpha$; \textit{Block Sparse Regularization} - promotes and transfers sparsity from the teacher network.}
		\vspace{-.5cm}
\label{fig:kdvi}
\end{figure*}

The cambrian explosion of machine learning applications over the past decade is largely due to deep neural networks (DNN) contributing to dramatic  performance improvements in the domains of speech, vision and text. Miniaturization of devices (smartphones, drones, head-mounts etc.) and significant progress in augmented/virtual reality devices pose constraints on CPU/GPU, memory and battery life, making it harder to deploy DNN based models on them~\cite{cheng2018model}. To address these requirements, compressing DNN and accelerating their performance in such constrained environments is the need of the hour. Thus, our aim is to enable real-time on-device inference with little or no compromise in model accuracy. 

\noindent In the literature, there are several approaches to model compression such as parameter pruning and sharing \cite{han2015deep}, low rank factorization \cite{denton2014exploiting}, compact convolutional filters \cite{HowardZCKWWAA17}, and knowledge distillation (KD) \cite{ba2014deep}. We are interested in KD, where the central idea is to distil knowledge from a large, complex and possibly a pre-trained teacher model to a small student network, by using the class distributions of teacher network for training the student. KD based approaches are attractive since it saves on retraining effort with respect to the teacher, and still leads to a smaller and a compressed student.  In \cite{bucilu2006model}, KD was first proposed for shallow models which was later extended to deep models in  \cite{hinton2015distilling}. Several variants of the KD approach have been proposed to achieve improved model compression such as \emph{FitNets} for wide and deep network compression \cite{romero2014fitnets}, \cite{luo2016face} for model compression in face recognition tasks, etc.

\noindent  A parallel approach to achieve sparsity in DNNs is by taking the Bayesian route. Bayesian neural networks (BNN) are robust to overfitting, they learn from small datasets and offer uncertainty estimates through the parameters of per-weight probability distributions. Furthermore, variational inference (VI) formulations can lead to clear separation between prediction accuracy and model complexity aiding in both, analysis and optimisation of DNNs \cite{NIPS2011_4329}, thus, contributing to explainable AI methods. One of the earliest contributions in the context of Bayesian inference for neural networks is the variational dropout (VD) technique which was proposed to infer the posterior of network weights, with a goal of learning these weights jointly with the dropout rate \cite{kingma2015variational} . In \cite{molchanov2017variational}, the authors proposed the sparse variational dropout (SVD) technique where they provided a novel approximation of the KL-divergence term in the VD objective \cite{kingma2015variational}, and showed that this leads to sparse weight matrices in fully-connected and convolutional layers. The authors in \cite{liu2018variational} proposed the variational Bayesian dropout (VBD) technique where, in addition to the prior on the weights, a hyperprior is assumed on the parameters of prior distribution. The authors in \cite{louizos2017bayesian} propose a technique to achieve compression beyond sparsity using a fixed point precision based encoding of weights.  While the sparsifying nature of these Bayesian inference technique was well-accepted, the authors in \cite{hron2018variational} showed that log-uniform prior employed in the above works led to an improper prior. Hence, they studied the objective from a non-Bayesian perspective, and  reinterpret the variational optimization as a type of penalised maximum likelihood estimation, with KL divergence as a regularizer.

\noindent In this work, we consider a BNN based student in a vanilla KD framework, where the student employs a variational penalized least squares objective function. The advantage of such an approach is two-fold: the student network is compact as compared to the teacher network by the virtue of KD, and in addition, the variational nature of the student objective allows us to employ several sparsity exploiting techniques such as SVD or VBD, hence achieving a sparse student. In particular, we expect that the hint from the teacher helps to retain the accuracy as achieved by the teacher, and yet obtain a compact and sparse student network. 

\noindent Block sparse constraints have been employed for realizing sparse neural networks~\cite{wen2016learning,wen2017learning}. We explore the ability of BSR to induce sparsity in the weights of student network, using the weights of the teacher network in the KD framework, since student and teacher networks are employed for related tasks. To the best of our knowledge, BSR has not been explored for inducing sparsity in the context of KD or in the Bayesian framework.

\noindent Experimentally, we compare our technique against the state-of-the-art network compression techniques (both Bayesian and non-Bayesian) in the literature such as \cite{han2015learning,balan2015bayesian,louizos2017bayesian}, and we demonstrate that our proposed technique strikes the right balance with respect to the tradeoff between compression and performance of a model. Furthermore, our proposed technique performs better in terms of accuracy, as compared to other Bayesian techniques such as \cite{balan2015bayesian,welling2011bayesian} in the low data regime, i.e, with fewer number of samples for training. Hence, teachers hint acts as a proxy for input data, leading to better accuracy as compared to scenarios where teacher hint is not available, for instance \cite{welling2011bayesian}.


\section{Preliminaries}
\label{sec:prilim}

\noindent In this section, we briefly describe the KD framework and VI techniques used for learning sparser networks. In the sequel, we consider a dataset consisting of $N$ samples, $\mathcal{D} = (x_n, y_n)^N_{n=1}$ for training.

\subsection{Knowledge Distillation}

\noindent In the \textit{Knowledge distillation} (KD) framework \cite{bucilu2006model,hinton2015distilling}, relevant information is transferred from a complex deeper network or ensemble of networks, called teacher network(s), to a simpler shallow network, called student network. Thus during inference, we obtain a compressed network consisting of fewer parameters, with minimal compromise on accuracy. The loss function, $\mathcal{L}_{KD}$, used for training the student MLP in KD framework is as follows:
\begin{align}
&\mathcal{L}_{KD}(\mathbf{x},\mathbf{y},\mathbf{W}_s,\mathbf{W}_t) = \mathcal{L}_S(\mathbf{y},\mathbf{z}^s) + \lambda_T ~\mathcal{L}_H(\mathbf{y}, \mathbf{z}^s, \mathbf{z}^t),\nonumber\\
&\qquad\mathbf{z}^s = f_s (\mathbf{x},\mathbf{y};\mathbf{W}_s), \quad \mathbf{z}^t = f_t (\mathbf{x},\mathbf{y};\mathbf{W}_t), \label{eq:kd_loss}
\end{align}
\noindent where $\mathbf{x} = [x_1,\hdots,x_N]$ and $\mathbf{y} = [y_1, \hdots, y_N]$ are the inputs and their corresponding labels, respectively, and $\lambda_T$ is a Langrangian multiplier. Further, $\mathbf{W}_s = \{ \mathbf{W}_s^{(l)}, 1 \leq l \leq L_s\} $, $\mathbf{W}_t = \{ \mathbf{W}_t^{(l)}, 1 \leq l \leq L_t\}$, where $\mathbf{W}_t^{(l)} \in \mathbb{R}^{K^{(l)}_t \times H^{(l)}_t}$, $\mathbf{W}_s^{(l)} \in \mathbb{R}^{K^{(l)}_s \times H^{(l)}_s}$   are the weight tensors of student and teacher networks, respectively. Note that,  $L_s$ and $L_t$ represent the number of layers in the student and teacher networks, respectively. The functions, $f_s(\cdot,\cdot;\cdot)$ and $f_t(\cdot,\cdot;\cdot)$ represent the student and teacher models that generate the respective logits $\mathbf{z}^s$ and $\mathbf{z}^t$. Further, the term $\mathcal{L}_{S}(\cdot,\cdot)$ represents the loss function associated to the student and  $\mathcal{L}_{H}(\cdot,\cdot,\cdot)$ represents the hint obtained from the teacher. In particular, the term $\mathcal{L}_{H}(\cdot,\cdot,\cdot)$ minimizes the differences in the outputs of both the networks and helps the student to mimic the teacher network.

\noindent Note that this analysis is performed on an MLP network and it can easily be extended to a CNN where student and teacher network weights are 4D-tensors, i.e, $\mathbf{W}_t^{(l)} \in \mathbb{R}^{K^{(l)}_t \times H^{(l)}_t \times C^{(l)}_t \times M^{(l)}_t}$ and $\mathbf{W}_s^{(l)} \in \mathbb{R}^{K^{(l)}_s \times H^{(l)}_s \times C^{(l)}_s \times M^{(l)}_s}$.

\subsection{Sparsity Through Variational Inference}

Consider a BNN with weights $\mathbf{W}$, and a prior distribution over the weights, $p(\mathbf{W})$. It has been shown that training a BNN involves optimizing a variational lower bound given by
\begin{equation}
\max_{\phi} ~ \mathcal{L}_\mathcal{D}(\phi) - D_{KL}(q_\phi(\mathbf{W}) \Vert p(\mathbf{W})),
\label{eq:elbo}
\end{equation}
\noindent where $q_\phi(\mathbf{W})$ is an approximation of the true posterior distribution of the weights of the network and $D_{KL}(q_\phi(\mathbf{W}) \Vert p(\mathbf{W}))$ is the KL-divergence between the true posterior and its approximation. The  expected log-likelihood, $\mathcal{L}_\mathcal{D}(\phi)$, is  given by
\begin{equation}
\mathcal{L}_\mathcal{D}(\phi) = \sum_{n=1}^N \mathbb{E}_{q_\phi(\mathbf{W})}[\log(p(y_n|x_n,\mathbf{W}))].
\end{equation}

\noindent It is evident from the above that based on different assumptions on the prior, $p(\mathbf{W})$, and approximate distribution, $q_{\phi}(\mathbf{W})$, it is possible to obtain different variational Bayesian formulations. Among such formulations, a sparsity promoting Bayesian inference method is the \textit{Sparse Variational Dropout} (SVD)~\cite{molchanov2017variational} technique. SVD assumes an improper log-scale uniform prior on the entries of the weight matrix  $\mathbf{W} \in \mathbb{R}^{K \times H}$, and $q_{\phi}(\mathbf{W})$ is derived to be a conditionally Gaussian distribution. Since SVD is based on  the VD technique \cite{kingma2015variational}, the corresponding BNN training involves learning the per-weight variational dropout parameter $\alpha_{k,h}$, and a parameter $\theta_{k,h}$ which parameterizes the distribution of the  weight $w_{k,h}$, i.e., the variational parameters are $\phi_{k,h} = [\alpha_{k,h},\theta_{k,h}], k \in \{1,\hdots,K\},h \in \{1, \hdots, H\}$. Further, VBD~\cite{liu2018variational} is an extension of the SVD technique, maintaining a log-scale uniform prior  on the entries of the weight matrix  $\mathbf{W}$. In addition, VBD employs a hierarchical prior on the parameters of the weight distributions which is consistent with the optimization of $\mathcal{L}_\mathcal{D}(\phi)$. In this work, we train the student network using the SVD and the VBD concepts in the loss function, and demonstrate the relative merits and demerits of the techniques. 

\section{Variational Student}
\label{sec:proposedMethod}
In this section, we state the main results of this work, which is a novel student-teacher architecture in the KD framework, where the student is a BNN. We also describe the novel, sparsity inducing loss functions that we employ to train the student.

\subsection{Training the Student Networks}

In Fig.~\ref{fig:kdvi}, we provide an overview of the training procedure of our proposed algorithm.  In accordance with the training technique used in the KD framework, we first train a teacher network and store its weights. We use these weights for generating the required hint during student network's training.

\noindent The student network is trained using loss function in \eqref{eq:kd_loss}, where
\begin{align}
\mathcal{L}_{S}(\mathbf{y},\mathbf{z^s}) &= -\frac{1}{N} \sum_{n=1}^N \mathbf{y}_n \log(\mathbf{z}^s_n)\\
\mathcal{L}_{H}(\mathbf{y},\mathbf{z}^s,\mathbf{z}^t) &= 2T^2 D_{KL} \left(\sigma'\left(\frac{\mathbf{z}^s}{T}\right) \Vert \sigma'\left(\frac{\mathbf{z}^t}{T}\right)\right).
\end{align}

\noindent In the above equations, $\textbf{y}$ is a one hot encoding of the ground truth classes, $\mathbf{z}^s$ and $\mathbf{z}^t$ are the output logits from student and teacher networks, respectively, as given in \eqref{eq:kd_loss}. Note that, $\mathcal{L}_{S}$ is the cross-entropy loss over $N$ data samples,  $D_{KL}$ represents the KL-divergence and $\sigma'(\cdot)$ represents a softmax function. Further, $T$ is called the temperature parameter which controls the \emph{softness} of probablity distribution over classes and the coefficient is chosen in accordance to~\cite{hinton2015distilling}, where they use the coefficient to scale the magnitudes of gradients produced by the `soft targets'.


\noindent To enforce sparsity in the student network, we use both SVD and VBD formulations as a variational regularizer (VR) in the loss function. Note that the main difference in these formulations arise in the KL-divergence approximation to be used in \eqref{eq:elbo}. The approximation of KL-divergence term proposed for SVD is as follows,
\begin{equation}
\centering
\begin{gathered}
D_{KL}^{SVD}(q(w_{k,h}|\theta_{k,h}, \alpha_{k,h}) \Vert p(w_{k,h})) \approx \\
 k_1 \sigma(k_2 + k_3 \log~\alpha_{k,h}) - 0.5~\log(1 + \alpha_{k,h}^{-1}) - k_1,
\end{gathered}
\end{equation}
where  $k_1 = 0.63576$, $k_2 = 1.87320$, and $k_3 = 1.48695 $. Further, $\sigma(.)$ represents a sigmoid function, and $\theta_{k,h}$ parameterizes the probability distribution of $w_{k,h}$. Owing to the hierarchical design of prior, VBD reduces the KL-divergence term in variational lower bound in \eqref{eq:elbo} as
\begin{equation}
D^{VBD}_{KL}(q(\mathbf{W}) \Vert p(\mathbf{W}|\gamma)) = \sum_{k=1}^K \sum_{h=1}^H 0.5~\log(1 + \alpha_{k,h}^{-1})
\end{equation}

\noindent Incorporating the sparsity constraint through VR, we have the loss function as,
\begin{align}
\mathcal{L}(\mathbf{x}, \mathbf{y},\mathbf{W}_s,\mathbf{W}_t, \alpha) =  \mathcal{L}_{S}(\mathbf{y},\mathbf{z}^s) + \lambda_T \mathcal{L}_{H}(\mathbf{y},\mathbf{z}^s,\mathbf{z}^t) \nonumber \\
+ \lambda_V \mathcal{L}_{KL}(\mathbf{W}_s, \alpha),
\label{equ:3termloss}
\end{align}
\noindent  where $\lambda_V$ is a regularization constant of KL-divergence term and $\mathcal{L}_{KL}$ could be $D_{KL}^{SVD}$ or $D_{KL}^{VBD}$ depending on the variant of the variational dropouts we use.

\noindent Furthermore, we pre-compute the output features from the teacher network corresponding to different input samples from the dataset and store them for reusing.  Hence, we can scale up our batch sizes, resulting in decrease in training time.

\subsection{Inducing Sparsity through Block Sparse Regularization}
\label{sec:BSR}

\noindent In this section, we describe the BSR constraint that we employ in framework described in the previous section. 

In \cite{hron2018variational}, the authors   reinterpreted the variational objective optimization as a type of penalised maximum likelihood estimation, with KL divergence as a regularizer. The non-Bayesian nature of the VI set-up provides us greater flexibility in incorporating constraints that may help in sparsity gains. Furthermore, the BSR constraint can also be viewed as a modification to the prior employed in the SVD and VBD techniques. The effectiveness of the BSR constraint in a multi-task learning (MTL) context is well-known \cite{argyriou2007multi}. Applying this constraint in our setup, we define $\mathbf{W}_{T:S}$ as the concatenation of $\mathbf{W}_t$ and $\mathbf{W}_s$ along the dimension of layers. We apply BSR on $\mathbf{W}_{T:S}$ and since the tasks performed by teacher and the student models are the same, we expect that this constraint promotes sparsity in the aggregated tensor. Since the teacher weights are fixed, only the student weights in $\mathbf{W}_{T:S}$ vary making it more sparse with training. Furthermore, in a typical scenario, the pre-trained teacher weights are non-sparse and this pushes the student weights to be sparser on applying BSR on $\mathbf{W}_{T:S}$. 

\noindent Let $M = \max_{l}(b(\mathbf{W}_i^{(l)})$ and $N = \max_{l}(h (\mathbf{W}_i^{(l)})$,  where $b(\cdot),h(\cdot)$ return the width and height of a weight matrix  and $1 \leq l \leq \max(L_t,L_s)$, and $i \in \{s,t\}$, i.e., $\mathbf{W}_{T:S}$ $\in \mathbb{R}^{M \times N \times L}$. We define BSR, termed as $\mathcal{R}_g(\cdot)$ as a function of $\mathbf{W}_{T:S}$, as
 
\begin{equation}
\mathcal{R}_g(\mathbf{W}_{T:S}) = \sum_{m=1}^{M} \left\lvert ~\sqrt[q]{{\sum_{n=1}^{N}}{\sum_{l=1}^{L}{\left( \mathbf{W}_{T:S}{(m,n,l)} \right) }^q} } ~\right\rvert.
\end{equation}

\noindent Note that the expression is a generic mixed norm of the form $l_1/l_q$. Specifically, a $l_1/l_\infty$ norm regulariser takes the following form:
\begin{equation}
\mathcal{R}_g(\mathbf{W}_{T:S}) = \sum_{m=1}^{M} \left\lvert ~\max_{n,l} \mathbf{W}_{T:S}{(m,n,l)} ~\right\rvert.
\label{equ:regGrpLasso}
\end{equation}

\noindent Similarly in case of CNNs, $\mathbf{W}_{T:S} \in \mathbb{R}^{M \times N \times K \times H \times L}$ is a 5D-tensor where $M,N,K,H,L$ take the maximum size in their respective dimension. Thus, in this case $\mathcal{R}_g(\mathbf{W}_{T:S})$ becomes,
\begin{equation}
\mathcal{R}_g(\mathbf{W}_{T:S}) = \sum_{m=1}^{M} \left\lvert ~\max_{n,k,h,l} \mathbf{W}_{T:S}{(m,n,k,h,l)} ~\right\rvert.
\end{equation} 

\noindent  We incorporate $\mathcal{R}_g(\mathbf{W}_{T:S})$ as a regularization term in \eqref{equ:3termloss} to arrive at the final loss function as given below:
\begin{align}
&\mathcal{L}(\mathbf{x}, \mathbf{y},\mathbf{W}_s,\mathbf{W}_t,\alpha) =  \mathcal{L}_{S}(\mathbf{y},\mathbf{z}^s) + \lambda_T \mathcal{L}_{H}(\mathbf{y},\mathbf{z}^s,\mathbf{z}^t)  \nonumber\\
&\qquad\qquad\qquad+ \lambda_V \mathcal{L}_{KL}(\mathbf{W}_s, \alpha) + \lambda_g \mathcal{R}_{g}(\mathbf{W}_{T:S}),
\label{equ:FinalLoss}
\end{align}
\noindent where $\lambda_g$ is an additional regularisation constant as compared to \eqref{equ:3termloss}. We use \eqref{equ:FinalLoss} to train the student network in our proposed method. Furthermore, in the next section, we compare the sparsity and accuracy of student networks trained with both $l_1/l_{\infty}$ and $l_1/l_2$ (group lasso) regularizations.


\section{Experimental Results and Discussions}
\label{sec:experiments}
\noindent In this section, we describe the experimental set up, evaluation criteria and the different experiments performed on different class of networks and datasets, followed by the experimental results.

\begin{table*}[!ht]
\begin{center}
\begin{tabular}{cccccc}
\hline
Network & \begin{tabular}{@{}c@{}}Test Error \\(in \%) \end{tabular}  & \begin{tabular}{@{}c@{}}Sparsity Per \\Layer (in \%) \end{tabular} & $R_s=\frac{\vert W \vert}{\vert W_{\neq 0} \vert}$ & \begin{tabular}{@{}c@{}}Compression Rate \\($R_c$) \end{tabular} & Parameters\\
\midrule \midrule
T1 & 1.59 & 0-0-0 & 1 & 1 & 2,395,210 \\
\textbf{Le-L-ST-SVD} & 1.94 & 86.08-95.14-95.90 & 7.77 & 9 & 266,200 \\
\textbf{Le-L-ST-VBD} & 1.88 & 91.37-96.39-95.20 & 12.43 & 9 & 266,200 \\
\midrule
IP~\cite{han2015learning} & 1.64 & 0-$5.8\times 10^{-6}$-0 & $1.4\times 10^{-6}$ & 7.52 & 318,145 \\
BDK~\cite{balan2015bayesian}$^*$ & 2.17 & 47.45-77.15-1.70 & 1.75 & 1.75 & 272,306 \\
BC~\cite{louizos2017bayesian} & 1.90 & 60.33-71.33-86.0 & 9.43 & 8.81 & 271,774 \\
\midrule
\end{tabular}
\end{center}
\caption{Comparison of different state-of-the-art Bayesian techniques against ours (in bold) on MNIST dataset with Le-L as student network. $^*$In case of BDK, the teacher network has $477,600$ parameters. }
\label{tab:le_eval}
\vspace{-.5cm}
\end{table*}

\noindent We use an 8 core Intel(R) Core(TM) i7-7820HK CPU, 32GB memory and an Nvidia(R) GeForce GTX 1080 GPU machine for experiments. The models are trained using PyTorch v0.4.1. For training and evaluation of MLPs and CNNs, we use popular datasets - MNIST~\cite{lecun2010mnist} and CIFAR-10~\cite{krizhevsky2009learning}. We split the training and testing data in the ratios of $6:1$ and $5:1$ for MNIST and CIFAR-10 datasets respectively. For all   experiments with MLP, we used Adam optimizer with a learning rate of $10^{-3}$ for 100-150 epochs on the MNIST dataset. For all experiments with CNN, we used Adam optimzer with a learning rate of $5 \times 10^{-4}$ for 100-150 epochs on the CIFAR dataset. For tackling early pruning problem we use warm up techniques \cite{molchanov2017variational} and set the value of $\lambda_V$. We use $\lambda_T=2$ and $\lambda_g=0.01$ in all the experiments.

\noindent Throughout the paper, the representation $a-b-c$ of a network structure represents number of nodes in different layers of the network. For the MLP-based experiments, we employ teacher \textbf{T1} with structure $\rm{1200}-\rm{1200}$ and students \textbf{S1} and \textbf{Le-L} with structures $\rm{500}-\rm{50}$ and $\rm{LeNet-300-100}$, respectively. Further, for CNN-based experiments, we use teacher \textbf{TC1}  which is a $\rm{VGG-19}$ network. The student for the CNN teacher is \textbf{Le-C} with structure $\rm{LeNet-5-Caffe}$\footnote{A modified version of LeNet5 from \url{https://goo.gl/4yI3dL}}. \textit{Simple} refers to an independently trained network, \textit{D} refers to a network trained with binary dropouts rate of $0.4$, \textit{KD} is trained with hint from the teacher, and \textit{ST} refers to networks trained with BSR in KD framework.

\subsection{Evaluation Criteria}

\noindent  We evaluate the model compression performance and compare the networks using the following metrics:
\begin{enumerate}
    \item Compression ratio, $R_c$, which is defined as $R_c = \frac{p_b}{p_{ac}}$, where $p_b$ and $p_{ac}$ are the number of trainable parameters before and after the compression, respectively.
    \item Sparsity induced compression ratio, $R_s$ which is defined as, $R_s = \frac{|W|}{|W_{\neq 0}|}$, where $|W|$ and $|W_{\neq 0}|$ are the number of weights and the number of non-zero weights of the DNN, respectively.
    \item Per-layer-sparsity.
    \item Memory footprint compression.
    \item Accuracy, which is the top-$1$ error in case of classification.
    \item Inference time.
\end{enumerate}

\subsection{Network Compression and Sparsification}
In this section, we present the results on network compression and sparsification for the neural network models and datasets specified in the previous section.

\noindent In Table~\ref{tab:le_eval}, we compare our method against the state-of-the-art Bayesian methods such as Bayesian Dark Knowledge (BDK)~\cite{balan2015bayesian}, and Bayesian Compression (BC)~\cite{louizos2017bayesian}. The methods proposed in our work provide better accuracy and sparsity as compared to Bayesian inference based approaches such as BDK and BC. In BDK, an untrained shallow teacher is simultaneously trained with the student. However, it is to be noted that BDK does not optimize on sparsity explicitly in their approach resulting in a lesser sparse solution. On the other hand, BC employs sparsity inducing priors in a Bayesian inference framework similar to SVD and VBD. The half-Cauchy distribution used as a prior in BC helps in group pruning, and hence provides higher degrees of sparsity as compared to SVD. However, when we use VBD, both the sparsity and the accuracy  of the student model increase with an additional aid of hints from the teacher network.

\noindent In addition, we observe a significant increase in training time for students in the BDK framework, even when we use similar parameter settings(learning rate and batch size) as given in BDK. In our work, the student MLP network takes about $0.5-1$ hours, while BDK takes about $2-2.5$ hours to train on the MNIST dataset over a same number of epochs. 

\subsubsection{Effects of Variational Inference}
\begin{figure*}[h!]
	\begin{center}
		\includegraphics[scale=.3]{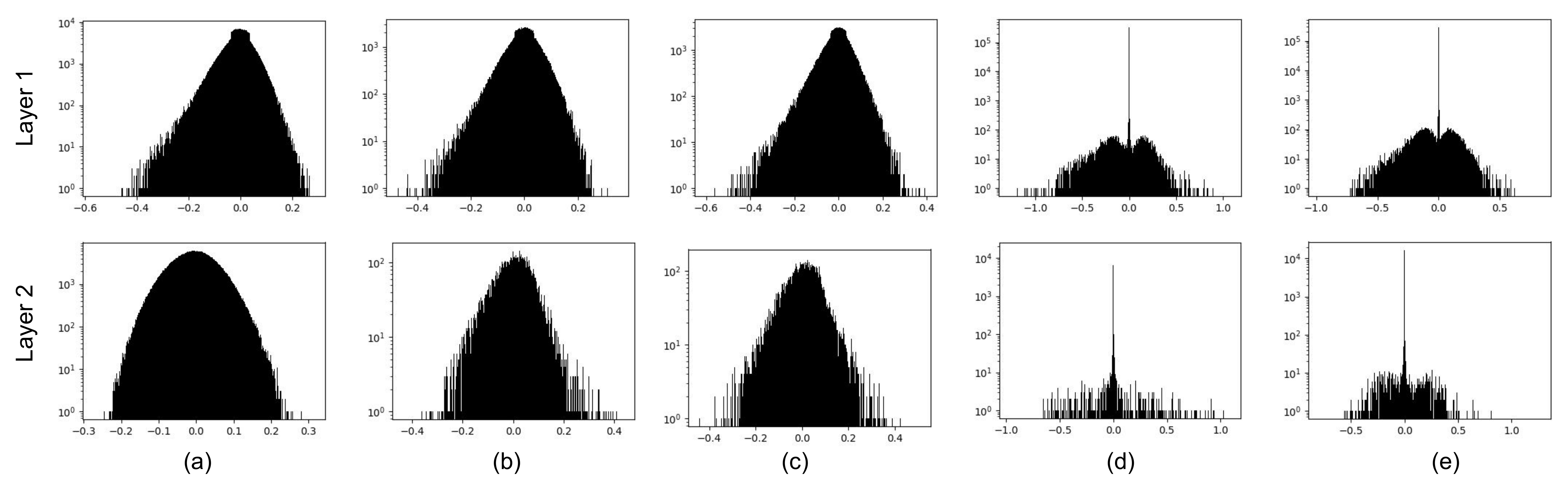}
	\end{center}
	\vspace{-.5cm}
		\caption{VI induces sparsity on student weight distributions. The figure depicts the weight distribution (y-axis is in log scale) of different networks. Subplots (a) and (b) represents weights of teacher and student networks when trained independently, (c) depicts the weights of the student network trained with teacher network's hint, (d) and (e) depict the weights of the \emph{Variational Student} using SVD, VBD, respectively and trained in a KD framework. We note that there is significant number of weights concentrated around $0$.}
\label{fig:dist_var}
\vspace{-.5cm}
\end{figure*}

We depict the effect of variational methods into the KD framework in Fig.~\ref{fig:dist_var} and Table~\ref{tab:mnist_eval}. Specifically, we show that the introduction of VI induces sparsity by a factor of $8\times$ to $17\times$. First of all, we observe that when teacher and student networks are trained independently, they learn weights which are non-sparse. Further, when the student network is trained with hint from the teacher network, it learns weights with negligible increase in sparsity, but a significant increase in accuracy. As expected, we obtain a drastic increase in sparsity when we apply SVD and VBD on student network, and retain accuracy performance as in the KD framework.

\subsubsection{Multi-Layered Perceptrons on MNIST}

\begin{table}[!ht]
\centering
\begin{tabular}{cccc}
\midrule
Network & \begin{tabular}{@{}c@{}}Test Error \\(in \%) \end{tabular}  & \begin{tabular}{@{}c@{}}Sparsity Per \\Layer (in \%) \end{tabular} & $R_s=\frac{\vert W \vert}{\vert W_{\neq 0} \vert}$\\
\midrule \midrule
T1 & 1.59 & $0.0001-0$ & 1 \\
S1-simple & 2.21 & 0-0 & 1 \\
S1-D & 1.69 & $0.00025-0$ & 1 \\
S1-KD-simple & 1.78 & $0.00025-0$ & 1 \\
S1-KD-D & 1.88 & $0.00076-0$ & 1 \\
\midrule
S1-SVD & 1.79 & 91.29-94.24 & 11.58 \\
S1-VBD & 1.84 & 93.36-96.02 & 15.18 \\
S1-KD-SVD & 1.75 & 92.32-94.98 & 13.11 \\
S1-KD-VBD & 1.72 & 88.05-90.22 & 8.47 \\
\textbf{S1-ST-SVD} & 1.81 & \textbf{94.07}-\textbf{96.64} & \textbf{17.31} \\
\textbf{S1-ST-VBD} & \textbf{1.67} & 93.83-96.53 & 16.65 \\
\hline
\end{tabular}
\caption{Evaluation of different MLPs on MNIST dataset. S1 family of student networks have $418,060$ parameters and give a compression of $\mathbf{47.42\times}$. Note: Using KD in conjunction with SVD/VBD does not increase the sparsity, but it improves the accuracy. To improve the performance gains in accuracy and sparsity, we employ the KD, SVD/VBD with BSR as shown in rows 6-11. }
\label{tab:mnist_eval}
\vspace{-.5cm}
\end{table}

We trained the MLP with our proposed algorithm on the MNIST dataset. These networks are trained with random initializations without any data augmentation. To compare the sparsity exploiting performance, we compare our method with VD \cite{molchanov2017variational} and VBD \cite{liu2018variational}, when used with KD framework. In Table~\ref{tab:mnist_eval}, we show compare the compression and sparsity performance achieved by our method as compared to the other variants in the KD framework. 

\begin{figure}[!ht]
	\begin{center}
		\includegraphics[width=.45\textwidth]{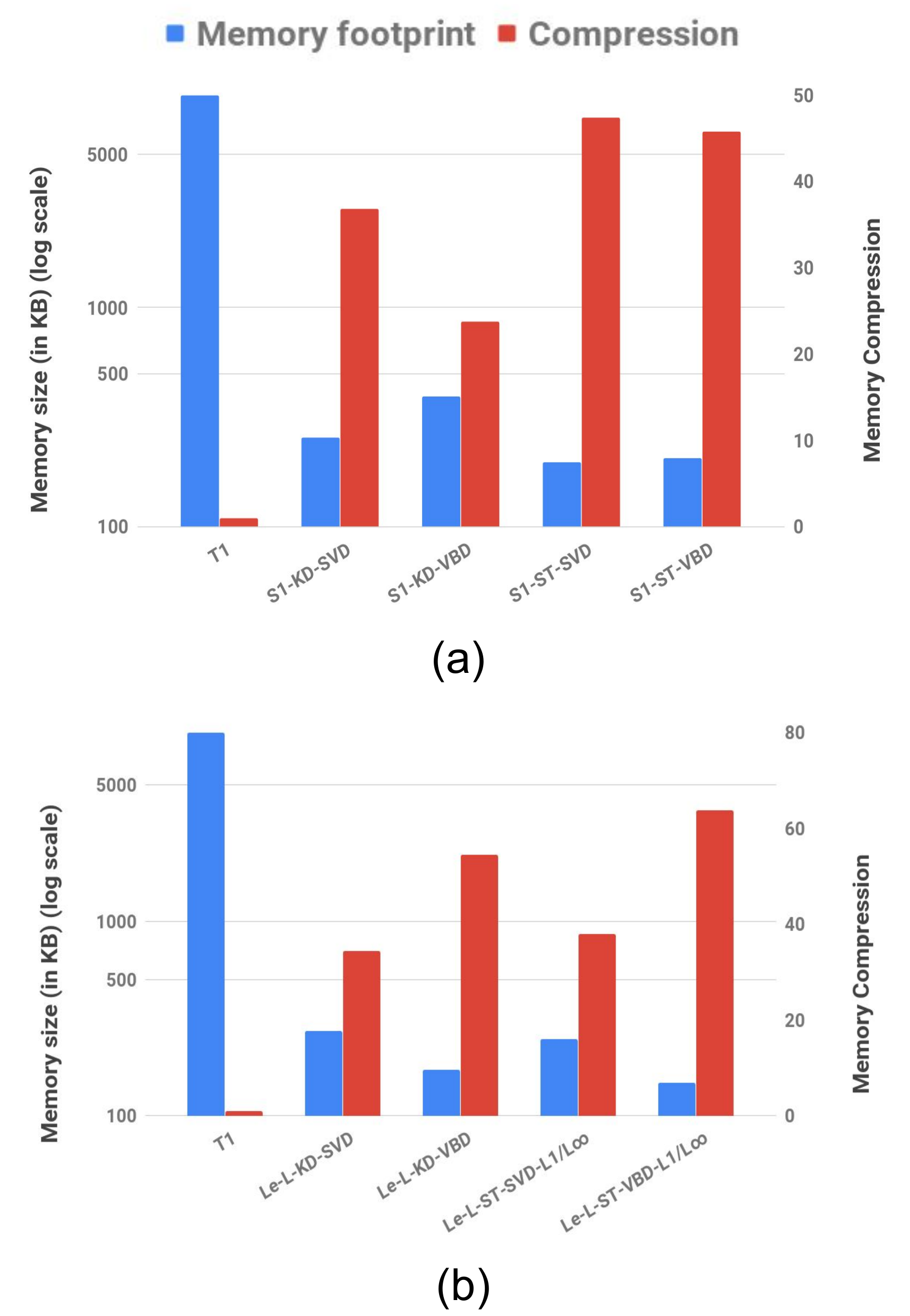}
	\end{center}
	\vspace{-0.3cm}
		\caption{Comparison of memory footprints and resultant compression from \emph{Variational Student} approach on S1  (a) and Le-L (b) families of students, respectively. We used compressed sparse row (CSR) format for reporting the efficient memory footprints of the sparse matrices. The best memory footprint compression for S1 and Le-L are $\mathbf{47.42\times}$ and $\mathbf{63.64\times}$ respectively.}
\label{fig:mem_compr}
\end{figure}

\noindent We observe that the proposed methods, namely ST and STB, outperform the KD variants in terms of both sparsity and accuracy, owing to the sparsity induced by BSR in addition to VR. Note that VBD variants outperform SVD variants in terms of accuracy in all the cases, hence proving the efficacy of hierarchical priors over log-uniform priors, which restricts the regularization performance of SVD~\cite{liu2018variational}. Further, Fig.~\ref{fig:mem_compr} shows the memory footprints of different student models. Note that ST and STB variants outperform others in terms of compression owing to the higher sparsity induced by BSR.

\subsubsection{VGG 19 (CNN) as Teacher on CIFAR10}

From Table~\ref{tab:cnn_eval} we see that compression due to sparsity is marginally enhanced in CNNs, but the number of parameters are reduced significantly, leading to large  gains on memory footprints. The VGG19 teacher takes $532.52$ MB, whereas a compressed student takes only $2.50$ MB, thus, achieving a compression of $\sim 213\times$. Owing to the teacher's hint, the sparser student variants perform better than \textit{Simple} students. Moreover, the sparser student variants outperform both, \textit{KD-Simple} and \textit{KD-D} variant due to the regularization power of both VR and BSR.

\begin{table}
\centering
\begin{tabular}{cccc}
\hline
Network & \begin{tabular}{@{}c@{}}Test Error \\(in \%) \end{tabular}  & \begin{tabular}{@{}c@{}}Sparsity Per \\Layer (in \%) \end{tabular} & $R_s=\frac{\vert W \vert}{\vert W_{\neq 0} \vert}$\\
\midrule\midrule
\\
TC1 & 14.21 & $0^*$ & 1 \\
Simple & 27.32 & \begin{tabular}{@{}c@{}}$\{0-0-6.4$\\$-0\}\times 10^{-4}$\end{tabular} & 1 \\
KD-Simple & 23.13 & \begin{tabular}{@{}c@{}}$\{0-0-1.6$\\$-0\}\times 10^{-4}$\end{tabular} & 1 \\
KD-D & 27.20 & \begin{tabular}{@{}c@{}}0-0-0-0\end{tabular} & 1 \\
\midrule
KD-SVD & 22.82 & \begin{tabular}{@{}c@{}}4.73-3.02-\\30.25-34.60 \end{tabular} & 1.47 \\
KD-VBD & 22.69 & \begin{tabular}{@{}c@{}} 2.18-2.55-\\34.21-35.62\end{tabular} & 1.49 \\
ST-SVD-$l_1$/$l_\infty$ & 22.68 & \begin{tabular}{@{}c@{}} 3.13-2.38-\\33.61-3518\end{tabular} & 1.48 \\
ST-SVD-$l_1$/$l_2$ & 22.72 & \begin{tabular}{@{}c@{}}2.07-2.14-\\27.75-33.68\end{tabular} & 1.37 \\
ST-VBD-$l_1$/$l_\infty$ & 22.71 & \begin{tabular}{@{}c@{}}2.80-2.30-\\31.95-34.80\end{tabular} & 1.44 \\
ST-VBD-$l_1$/$l_2$ & 22.60 & \begin{tabular}{@{}c@{}}2.60-2.32-\\31.95-34.80\end{tabular} & 1.44 \\
\midrule
\end{tabular}
\caption{Evaluation of CNNs on CIFAR dataset. Le-C family of student networks have $657,080$ parameters and thus give a compression of $\mathbf{212.47\times}$. $^{*}$ Representative of all layers of VGG19.}
\label{tab:cnn_eval}
		\vspace{-.3cm}
\end{table}

\subsubsection{Effects of Block Sparse Regularization}
In Table~\ref{tab:mnist_eval} we demonstrate that applying the BSR constraint increases the sparsity with the variational techniques. However, in scenarios where spatial information is associated with the network weights, such as in CNNs, BSR performs relatively inferior to the cases with no spatial constraints on weights (refer to Table~ \ref{tab:cnn_eval}). Furthermore Fig. 4, shows that we obtain more zeroes on taking teacher network’s weights into consideration in $\mathbf{W}_{T:S}$ indicating its role in inducing sparsity in student.

\begin{figure}[h!]
	\begin{center}
		\includegraphics[width=.45\textwidth]{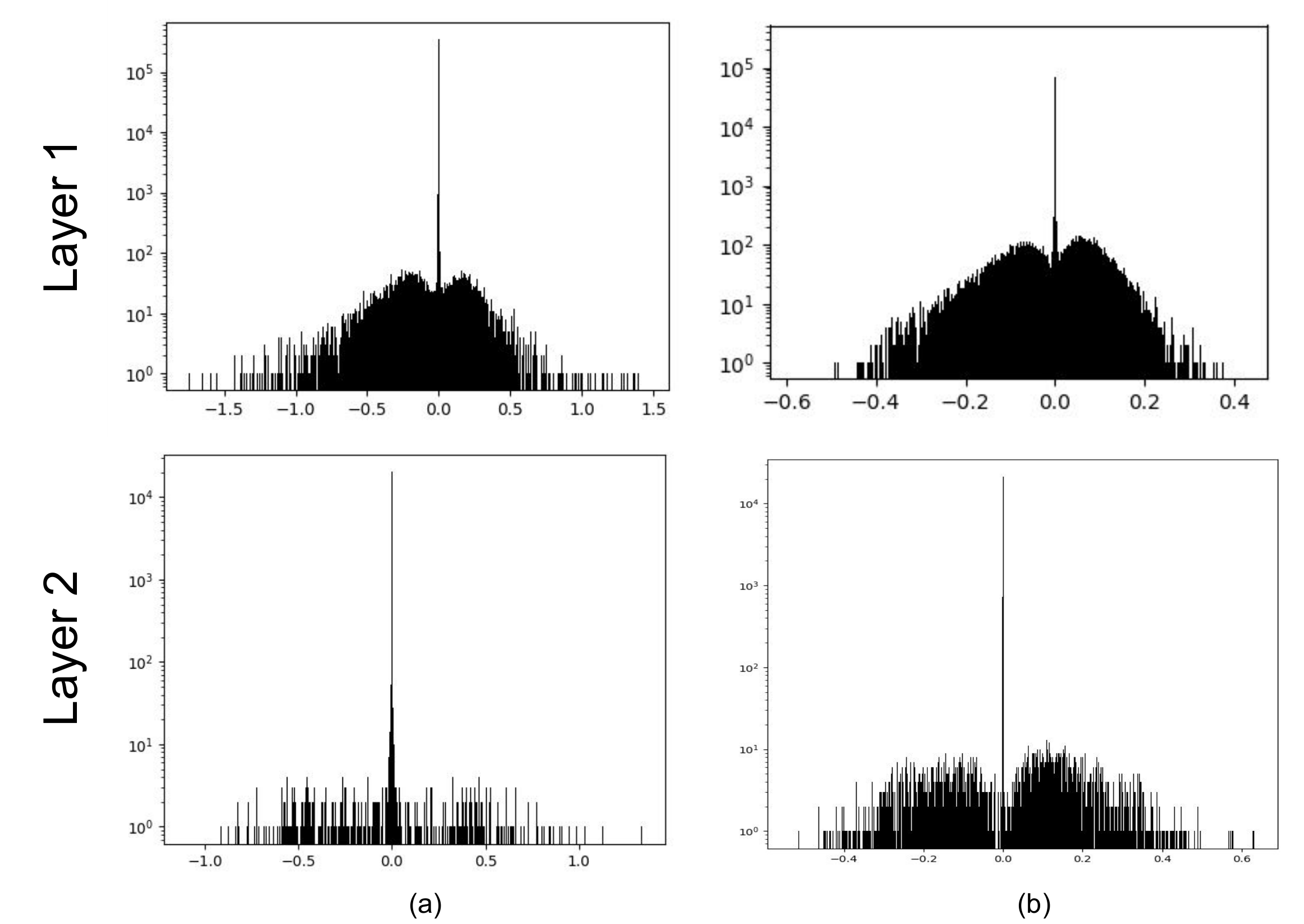}
	\end{center}
		\caption{Sparsity induced by BSR on student weight distributions. The figure shows resultant weight distribution (y-axis is in log scale) of a student MLP when (a) BSR is  applied on a composite tensor consisting of weights from both student and teacher networks, and (b) BSR is applied only on student's weights. Note the weights are concentrated around 0.}
\label{fig:dist_bsr}
\end{figure}

\subsubsection{Performance in Low Data Regime}

\begin{figure}
	\begin{center}
		\includegraphics[scale=0.22]{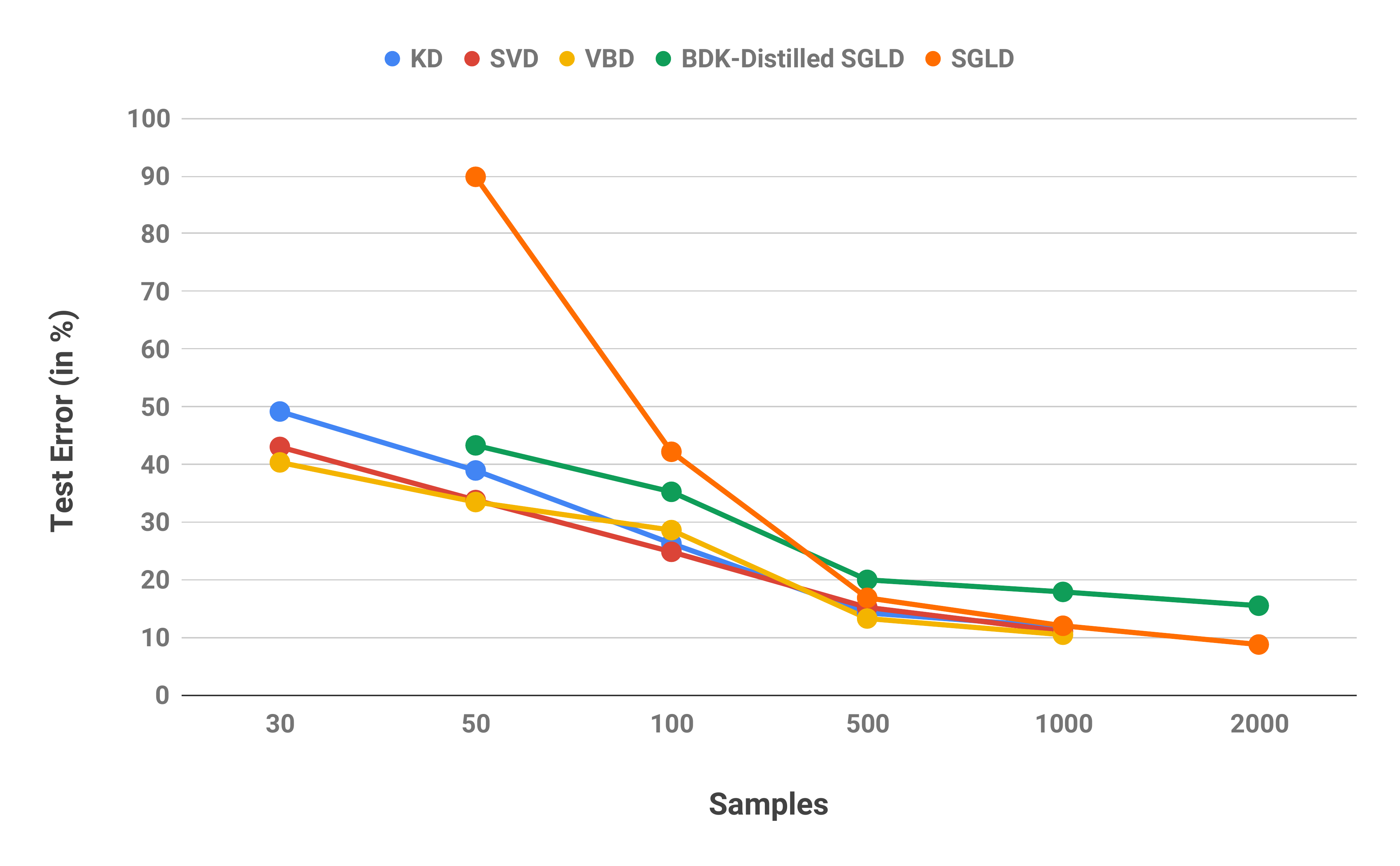}
	\end{center}
\caption{Performance evaluation of test error with varying number of training samples.}
\label{fig:lowdata}
\end{figure}

In Fig.~\ref{fig:lowdata}, we compare the performance of our proposed technique in training scenarios with less samples. We compare our method against BDK~\cite{balan2015bayesian} and SGLD~\cite{welling2011bayesian} which are Bayesian training approaches. The methods which are based on KD framework have relatively better performance than other techniques, such as SGLD, in low data scenario. Intuitively, the hint that we obtain from the teacher behaves as a proxy for observations in the input dataset.

\subsection{Discussion}
We acknowledge that overall posterior induced is a combination of a log-scale uniform prior\cite{kingma2015variational} and the BSR regulariser, and hence, an improper posterior. The authors \cite{hron2018variational} rigorously prove that the posteriors obtained in \cite{molchanov2017variational} renders the setup to be non-bayesian in the strict sense. They also discuss that albeit being non-Bayesian, the variational formulation provide good empirical results when the considered metrics are accuracy, etc. Since the goal of our paper is aligned with the objectives in \cite{molchanov2017variational}, we chose this technique to obtain a sparse student network. Furthermore, we empirically show that SVD \cite{molchanov2017variational} and VBD \cite{liu2018variational} provide good results in the KD framework, and hence, the improper nature of the posterior does not affect us much. 

\subsection{Runtime Analysis}

\begin{figure}[h!]
	\begin{center}
		\includegraphics[scale=.16]{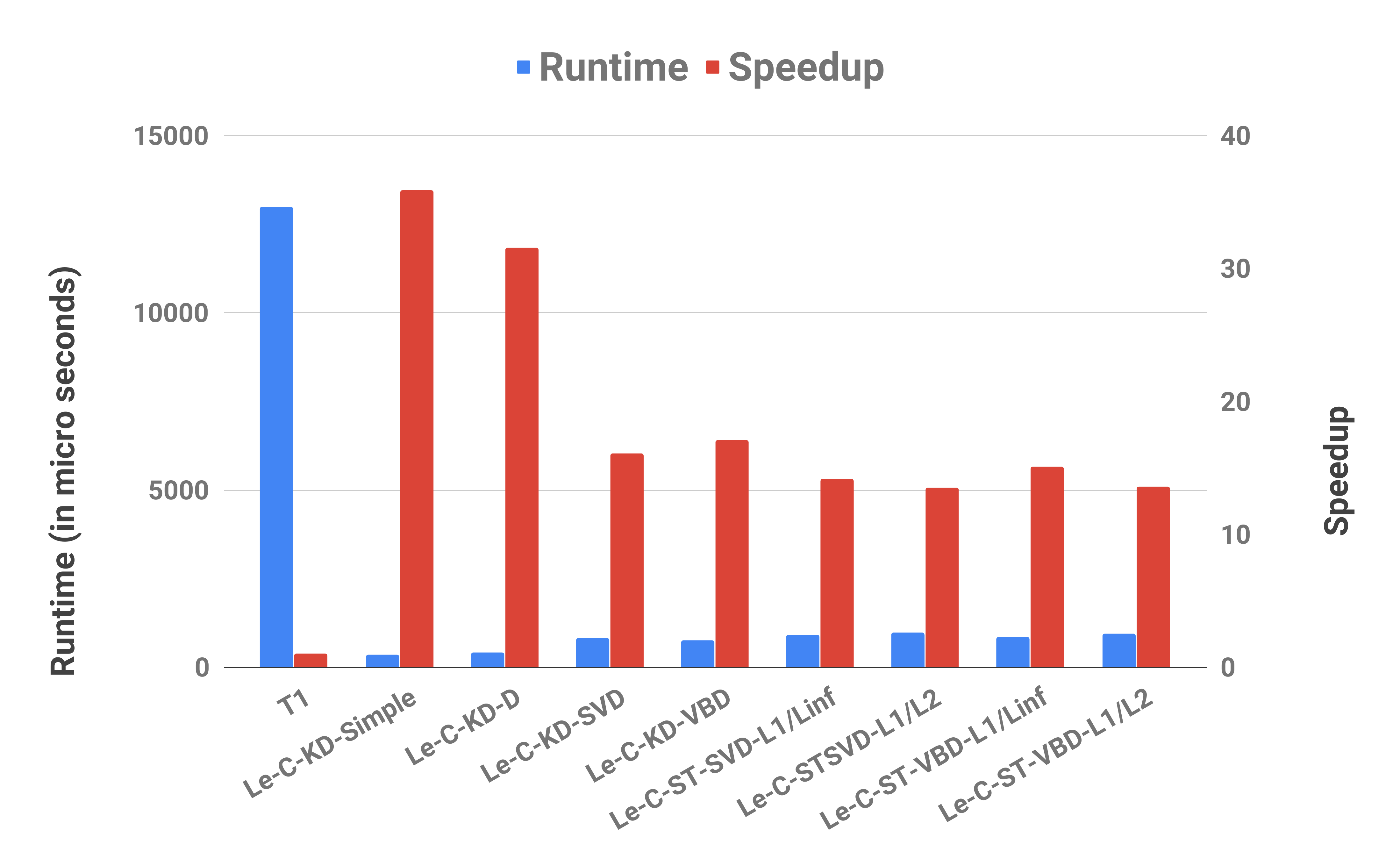}
	\end{center}
\caption{Speedup of different variants of CNNs.}
\label{fig:speedup_cnn}
\end{figure}

We note the inference time of teacher MLP to be $\rm0.29$ ms and the student variants have inference time in the range $0.257-0.470$ ms. We observe that for MLPs, the variational student and BSR variants have similar inference times.  Although both the variants have different computations to perform during training, but during inference they have same operations owing to similar student structures. We also notice that \textit{Simple} variants have lesser inference time compared to other variants as they avoid additional computation involving thresholding operation on $\alpha$ and multiplying the resultant mask with the weights. We see similar trends for CNNs as well in Fig.~\ref{fig:speedup_cnn}.

\vspace{-3mm}
\section{Conclusions}
\label{sec:conclusions}

In summary, we introduced \emph{Variational Student} that sparsifies the neural network through a combination of VI and BSR techniques in a KD framework. We demonstrate that our work compresses the memory footprints of MLPs and CNNs by factors of $\rm\approx64$ and $\rm\approx213$, with minimal increase in test error. We argue that VI based methods such as VBD and SVD techniques when employed in the student architecture in the KD framework contribute to compression and hence speed-up. Leveraging the effectiveness of KD framework, we demonstrate an improvement in accuracy in low-data regime as well. In future we would like to extend our approach to deeper CNN teacher networks applied to different domains, and port these models on smartphone.

\appendix

\bibliographystyle{named}
\bibliography{aaai}

\begin{thebibliography}{}

\bibitem[\protect\citeauthoryear{Argyriou \bgroup \em et al.\egroup
  }{2007}]{argyriou2007multi}
Andreas Argyriou, Theodoros Evgeniou, and Massimiliano Pontil.
\newblock Multi-task feature learning.
\newblock In {\em Advances in neural information processing systems}, pages
  41--48, 2007.

\bibitem[\protect\citeauthoryear{Ba and Caruana}{2014}]{ba2014deep}
Jimmy Ba and Rich Caruana.
\newblock Do deep nets really need to be deep?
\newblock In {\em Advances in neural information processing systems}, pages
  2654--2662, 2014.

\bibitem[\protect\citeauthoryear{Balan \bgroup \em et al.\egroup
  }{2015}]{balan2015bayesian}
Anoop~Korattikara Balan, Vivek Rathod, Kevin~P Murphy, and Max Welling.
\newblock Bayesian dark knowledge.
\newblock In {\em Advances in Neural Information Processing Systems}, pages
  3438--3446, 2015.

\bibitem[\protect\citeauthoryear{Buciluǎ \bgroup \em et al.\egroup
  }{2006}]{bucilu2006model}
Cristian Buciluǎ, Rich Caruana, and Alexandru Niculescu-Mizil.
\newblock Model compression.
\newblock In {\em Proceedings of the 12th ACM SIGKDD international conference
  on Knowledge discovery and data mining}, pages 535--541. ACM, 2006.

\bibitem[\protect\citeauthoryear{Cheng \bgroup \em et al.\egroup
  }{2018}]{cheng2018model}
Yu~Cheng, Duo Wang, Pan Zhou, and Tao Zhang.
\newblock Model compression and acceleration for deep neural networks: The
  principles, progress, and challenges.
\newblock {\em IEEE Signal Processing Magazine}, 35(1):126--136, 2018.

\bibitem[\protect\citeauthoryear{Denton \bgroup \em et al.\egroup
  }{2014}]{denton2014exploiting}
Emily~L Denton, Wojciech Zaremba, Joan Bruna, Yann LeCun, and Rob Fergus.
\newblock Exploiting linear structure within convolutional networks for
  efficient evaluation.
\newblock In {\em Advances in neural information processing systems}, pages
  1269--1277, 2014.

\bibitem[\protect\citeauthoryear{Graves}{2011}]{NIPS2011_4329}
Alex Graves.
\newblock Practical variational inference for neural networks.
\newblock In J.~Shawe-Taylor, R.~S. Zemel, P.~L. Bartlett, F.~Pereira, and
  K.~Q. Weinberger, editors, {\em Advances in Neural Information Processing
  Systems 24}, pages 2348--2356. Curran Associates, Inc., 2011.

\bibitem[\protect\citeauthoryear{Han \bgroup \em et al.\egroup
  }{2015}]{han2015learning}
Song Han, Jeff Pool, John Tran, and William Dally.
\newblock Learning both weights and connections for efficient neural network.
\newblock In {\em Advances in neural information processing systems}, pages
  1135--1143, 2015.

\bibitem[\protect\citeauthoryear{Han \bgroup \em et al.\egroup
  }{2016}]{han2015deep}
Song Han, Huizi Mao, and William~J Dally.
\newblock Deep compression: Compressing deep neural networks with pruning,
  trained quantization and huffman coding.
\newblock {\em International Conference on Learning Representations}, 2016.

\bibitem[\protect\citeauthoryear{Hinton \bgroup \em et al.\egroup
  }{2015}]{hinton2015distilling}
Geoffrey Hinton, Oriol Vinyals, and Jeff Dean.
\newblock Distilling the knowledge in a neural network.
\newblock {\em arXiv preprint arXiv:1503.02531}, 2015.

\bibitem[\protect\citeauthoryear{Howard \bgroup \em et al.\egroup
  }{2017}]{HowardZCKWWAA17}
Andrew~G. Howard, Menglong Zhu, Bo~Chen, Dmitry Kalenichenko, Weijun Wang,
  Tobias Weyand, Marco Andreetto, and Hartwig Adam.
\newblock Mobilenets: Efficient convolutional neural networks for mobile vision
  applications.
\newblock {\em CoRR}, abs/1704.04861, 2017.

\bibitem[\protect\citeauthoryear{Hron \bgroup \em et al.\egroup
  }{2018}]{hron2018variational}
Jiri Hron, Alexander G de~G Matthews, and Zoubin Ghahramani.
\newblock Variational bayesian dropout: pitfalls and fixes.
\newblock {\em arXiv preprint arXiv:1807.01969}, 2018.

\bibitem[\protect\citeauthoryear{Kingma \bgroup \em et al.\egroup
  }{2015}]{kingma2015variational}
Durk~P Kingma, Tim Salimans, and Max Welling.
\newblock Variational dropout and the local reparameterization trick.
\newblock In {\em Advances in Neural Information Processing Systems}, pages
  2575--2583, 2015.

\bibitem[\protect\citeauthoryear{Krizhevsky and
  Hinton}{2009}]{krizhevsky2009learning}
Alex Krizhevsky and Geoffrey Hinton.
\newblock Learning multiple layers of features from tiny images.
\newblock 2009.

\bibitem[\protect\citeauthoryear{LeCun \bgroup \em et al.\egroup
  }{2010}]{lecun2010mnist}
Yann LeCun, Corinna Cortes, and CJ~Burges.
\newblock Mnist handwritten digit database.
\newblock {\em AT\&T Labs [Online]. Available: http://yann. lecun.
  com/exdb/mnist}, 2, 2010.

\bibitem[\protect\citeauthoryear{Liu \bgroup \em et al.\egroup
  }{2018}]{liu2018variational}
Yuhang Liu, Wenyong Dong, Lei Zhang, Dong Gong, and Qinfeng Shi.
\newblock Variational bayesian dropout.
\newblock {\em arXiv preprint arXiv:1811.07533}, 2018.

\bibitem[\protect\citeauthoryear{Louizos \bgroup \em et al.\egroup
  }{2017}]{louizos2017bayesian}
Christos Louizos, Karen Ullrich, and Max Welling.
\newblock Bayesian compression for deep learning.
\newblock In {\em Advances in Neural Information Processing Systems}, pages
  3288--3298, 2017.

\bibitem[\protect\citeauthoryear{Luo \bgroup \em et al.\egroup
  }{2016}]{luo2016face}
Ping Luo, Zhenyao Zhu, Ziwei Liu, Xiaogang Wang, and Xiaoou Tang.
\newblock Face model compression by distilling knowledge from neurons.
\newblock In {\em Thirtieth AAAI Conference on Artificial Intelligence}, 2016.

\bibitem[\protect\citeauthoryear{Molchanov \bgroup \em et al.\egroup
  }{2017}]{molchanov2017variational}
Dmitry Molchanov, Arsenii Ashukha, and Dmitry Vetrov.
\newblock Variational dropout sparsifies deep neural networks.
\newblock In {\em International Conference on Machine Learning}, pages
  2498--2507, 2017.

\bibitem[\protect\citeauthoryear{Romero \bgroup \em et al.\egroup
  }{2014}]{romero2014fitnets}
Adriana Romero, Nicolas Ballas, Samira~Ebrahimi Kahou, Antoine Chassang, Carlo
  Gatta, and Yoshua Bengio.
\newblock Fitnets: Hints for thin deep nets.
\newblock {\em arXiv preprint arXiv:1412.6550}, 2014.

\bibitem[\protect\citeauthoryear{Welling and Teh}{2011}]{welling2011bayesian}
Max Welling and Yee~W Teh.
\newblock Bayesian learning via stochastic gradient langevin dynamics.
\newblock In {\em Proceedings of the 28th international conference on machine
  learning (ICML-11)}, pages 681--688, 2011.

\bibitem[\protect\citeauthoryear{Wen \bgroup \em et al.\egroup
  }{2016}]{wen2016learning}
Wei Wen, Chunpeng Wu, Yandan Wang, Yiran Chen, and Hai Li.
\newblock Learning structured sparsity in deep neural networks.
\newblock In {\em Advances in neural information processing systems}, pages
  2074--2082, 2016.

\bibitem[\protect\citeauthoryear{Wen \bgroup \em et al.\egroup
  }{2017}]{wen2017learning}
Wei Wen, Yuxiong He, Samyam Rajbhandari, Minjia Zhang, Wenhan Wang, Fang Liu,
  Bin Hu, Yiran Chen, and Hai Li.
\newblock Learning intrinsic sparse structures within long short-term memory.
\newblock {\em arXiv preprint arXiv:1709.05027}, 2017.

\end{thebibliography}

\end{document}